\documentclass[10pt,twocolumn,letterpaper]{IEEEtran}
\usepackage[utf8]{inputenc}

\usepackage{amsmath, amssymb, amsfonts}
\usepackage{graphicx}
\usepackage{cite}
\usepackage{hyperref}

\usepackage[nameinlink]{cleveref}
\crefname{figure}{Fig.}{Figs.}
\Crefname{figure}{Fig.}{Figs.}

\usepackage{booktabs}
\usepackage[caption=false,font=footnotesize]{subfig} % IEEEtran recommended

\usepackage{siunitx}
\usepackage{tabularx}
\usepackage{xcolor}
\usepackage{soul}

\definecolor{hlY}{RGB}{255,249,196} % soft yellow
\definecolor{hlG}{RGB}{200,230,201} % soft green
\definecolor{hlB}{RGB}{187,222,251} % soft blue
\definecolor{hlP}{RGB}{225,190,231} % soft purple (lavender)

\sethlcolor{yellow}
\soulregister\cite7
\soulregister\ref7
\soulregister\cref7
\soulregister\Cref7

\usepackage{balance}

\title{Learning to Tune Pure Pursuit in Autonomous Racing: Joint Lookahead and Steering-Gain Control with PPO}

\author{Mohamed Elgouhary and Amr S.~El-Wakeel%
\thanks{The authors are with the Lane Department of Computer Science and Electrical Engineering, West Virginia University, Morgantown, WV, USA
(\texttt{\{mae00018@mix.wvu.edu, amr.elwakeel@mail.wvu.edu}).}%
}

\begin{document}

%\markboth{IEEE ROBOTICS AND AUTOMATION LETTERS. PREPRINT VERSION. ACCEPTED \textbf{FEBRUARY, 2026}}%
%{Elgouhary \MakeLowercase{\textit{et al.}}: Learning to Tune Pure Pursuit in Autonomous Racing}

\maketitle

\begin{abstract}
Pure Pursuit (PP) is widely used in autonomous racing for real-time path tracking due to its efficiency and geometric clarity, yet performance is highly sensitive to how key parameters—lookahead distance and steering gain—are chosen.
Standard velocity-based schedules adjust these only approximately and often fail to transfer across tracks and speed profiles.
We propose a reinforcement-learning (RL) approach that \emph{jointly} chooses the lookahead \(L_d\) and a steering gain \(g\) online using Proximal Policy Optimization (PPO).
The policy observes compact state features (speed and curvature taps) and outputs \((L_d,g)\) at each control step.
Trained in F1TENTH Gym and deployed in a ROS~2 stack, the policy drives PP directly (with light smoothing) and requires no per-map retuning.
Across simulation and real-car tests, the proposed RL--PP controller that jointly selects \((L_d,g)\)
consistently outperforms fixed-lookahead PP, velocity-scheduled adaptive PP, and an RL lookahead-only variant,
and it also exceeds a kinematic MPC raceline tracker under our evaluated settings in lap time, path-tracking accuracy, and steering smoothness, demonstrating that policy-guided parameter tuning can reliably improve classical geometry-based control.
\end{abstract}


\section{Introduction}

Intelligent and connected vehicles must track reference paths accurately across widely varying speeds and curvatures while preserving stability and comfort. Geometric controllers remain attractive in this setting for their transparency and low computational cost, with Pure Pursuit (PP)~\cite{Pure-Pursuit} directing the vehicle toward a lookahead point and converting geometry into a steering command suitable for real-time use.

A longstanding limitation of PP is its sensitivity to important parameters---most notably the lookahead distance and the effective steering gain. Small lookahead values yield agile cornering but can excite oscillations on straights; large values smooth the motion yet understeer in tight bends. Likewise, an overly aggressive steering gain can induce noise, whereas a conservative gain slows convergence. Rule-based schedules that scale lookahead (and sometimes gain) with velocity or curvature partially address these effects, but their fixed functional forms and tuned coefficients often fail to transfer across tracks, speed profiles, and platforms. Model-predictive control (MPC) can achieve strong performance, but typically requires more modeling effort and online optimization.
Instead of replacing PP, we keep its geometric steering law and learn to tune a small set of parameters to improve performance while preserving simplicity and interpretability.
We propose a data-driven alternative in which a reinforcement learning (RL) policy \emph{jointly} sequences the lookahead distance $L_d$ and a steering gain $g$ online.
We formulate parameter sequencing as a continuous control problem and train with Proximal Policy Optimization (PPO)~\cite{PPO} to map compact state features---vehicle speed and curvature taps at near/mid/far horizons---into a 2-D action $(L_d, g)$. The policy is trained in the F1TENTH Gym~\cite{f1tenth} using a minimum-curvature raceline~\cite{tum} to supply waypoints and curvature; optimization is stabilized through KL-constrained PPO updates, observation/return normalization, gradient clipping, and a scheduled learning rate chosen via Optuna~\cite{Optuna}. In simulation, we increase task difficulty by uniformly scaling the raceline speed profile so that the policy must learn a curvature- and speed-aware schedule rather than overfitting to a single fixed speed profile.

The learned policy is integrated into a ROS~2 control stack that preserves PP's geometric steering law and exposes two lightweight interfaces (topics) for $(L_d, g)$. A first-order smoother mitigates sudden action changes, and a safety ``teacher'' (linear $v \!\mapsto\! L_d$ and $v \!\mapsto\! g$) serves as a fallback if RL commands become stale.

To focus on the effect of learned parameter tuning while preserving PP's interpretability and runtime efficiency,
We evaluate against multiple Pure Pursuit (PP) baselines---including fixed-lookahead and velocity-scheduled variants---as well as an MPC raceline tracker.
We evaluate in simulation (zero-shot across unseen tracks) and on the physical F1TENTH platform at speeds limited by safety and track size, and we monitor PPO diagnostics (approximate KL, clip fraction, action standard deviation, and
value loss as illustrated in Figs.~\ref{fig:ppo_tb_a} and \ref{fig:ppo_tb_b}) to verify stable optimization consistent with the chosen learning-rate settings.

The main contributions of this paper are:
\begin{itemize}
    \item An RL--Pure Pursuit framework that adapts the lookahead distance $L_d$ and steering gain $g$ online while preserving PP's simplicity and interpretability.

    \item A practical training recipe (reward design, hyperparameter tuning, and stability techniques) that learns a curvature--speed-aware $(L_d,g)$ policy with smooth, accurate tracking.

    \item A simulation and real-car evaluation in a ROS~2 F1TENTH stack, including ablations and comparisons to fixed PP, velocity-scheduled PP, and a kinematic MPC tracker, showing improved lap time, lateral error, and steering smoothness.

    \item An interpretability study that visualizes how $L_d$ and $g$ vary with track geometry and speed along the raceline.
\end{itemize}

This work shows how deep reinforcement learning can complement geometric control to improve adaptability and robustness while preserving interpretable, real-time path tracking.

\section{Related Work}
Path tracking is a core problem in autonomous driving, addressed by classical methods \cite{path-tracking} and learning-based approaches \cite{Learning-Based}. Among geometric controllers, Pure Pursuit (PP) \cite{Pure-Pursuit} and Stanley \cite{Stanley} remain popular due to their simplicity and real-time performance, including in racing settings \cite{racing}. A central limitation of PP is its sensitivity to the lookahead distance, which trades responsiveness against smoothness. Consequently, many works propose adaptive lookahead schemes based on speed and/or curvature \cite{Pure-Pursuit,curve2}, but still rely on hand-designed functional forms and tuned coefficients.

Model predictive control (MPC) optimizes over a finite horizon with dynamics and constraints and is widely used as a strong baseline in autonomous racing. While MPC can achieve precise tracking and constraint handling, it typically requires accurate modeling, online optimization, and careful tuning of costs and constraints, which may be challenging on resource-limited platforms.

Reinforcement learning (RL) provides a data-driven alternative for context-dependent control \cite{RL}, with methods such as DQN \cite{DQN}, DDPG \cite{DDPG}, and PPO \cite{PPO} applied across robotics and autonomous driving. End-to-end policies \cite{End-to-End} can be effective but often face interpretability and transfer challenges \cite{challenges,safety,sim}. To balance adaptability with structure, hybrid approaches use RL to tune parameters of classical controllers rather than replace them outright \cite{pid,gain,comparison}. However, most prior work focuses on throttle or full-actuation policies, with less emphasis on learning to tune the core geometric path-following parameters in PP.
While RL-driven lookahead adaptation has been explored in other domains (e.g., heavy trucks) \cite{truck},
we explicitly benchmark against an \emph{adaptive PP} baseline (a velocity-linear Pure Pursuit schedule) and include an MPC raceline tracker as a stronger model-based reference in both simulation and hardware experiments.

Finally, we use minimum-curvature racelines from TUM’s \texttt{global\_racetrajectory\_optimization} \cite{tum} as the shared reference for all controllers.

Prior adaptive PP methods adjust $L_d$ with hand-designed rules (speed/curvature/preview), while RL-based tuning has been used to adapt gains of classical controllers or to learn end-to-end tracking.
Our contribution differs in that we (i) keep the PP steering law fixed and interpretable, (ii) learn a \emph{joint} continuous schedule for both $L_d$ and a gain $g$ using compact curvature-preview features, and (iii) evaluate zero-shot generalization to unseen tracks and deploy within a ROS~2 stack on a real F1TENTH vehicle.

\section{Framework and Methodology}
% brief lead-in for the section
This section details the full stack used in our study.
We estimate pose with a LiDAR-based Monte Carlo localization (MCL) filter, generate a global reference using minimum-curvature optimization \cite{tum}, and couple a classical Pure Pursuit (PP) tracker with a learned policy that selects the lookahead distance and the steering gain online.
We then specify observation and action spaces, the reward, PPO training, and how these components integrate in simulation and on the real F1TENTH platform \cite{f1tenth} for fair comparison against multiple PP baselines (fixed-lookahead and velocity-scheduled variants) and an MPC raceline tracker.

From a controller-design perspective, our goal is to investigate how far a lightweight
geometric Pure Pursuit approach can be pushed by exposing only a small set of parameters to learning.
We therefore build our stack around PP and evaluate against standard tracking baselines under a common reference trajectory and localization stack.

\subsection{Localization and Global Reference}
We estimate pose using a standard LiDAR-based Monte Carlo localization (MCL) particle filter~\cite{pf} over a prebuilt occupancy grid. Particles are propagated with a kinematic motion model, reweighted by a beam-based scan likelihood using grid ray-casting, and resampled when the effective sample size drops; the state estimate is taken as the weighted mean (circular mean for heading).

For the global reference, we use TUM’s \texttt{global\_racetrajectory\_optimization}~\cite{tum} to compute a smooth minimum-curvature raceline within track boundaries. The optimizer returns a closed-loop trajectory with associated curvature $\kappa(s)$ and a friction-limited speed profile $v_{\max}(s)$, which we export as waypoints $\{x(s),y(s),\kappa(s),v_{\max}(s)\}$.
This globally smooth geometry provides both the tracking reference and curvature features used by the controller and the RL policy.

\begin{figure}[t]
  \centering
  \captionsetup{skip=2pt} % spacing only

  \includegraphics[
    page=1,
    width=\columnwidth,
    keepaspectratio,
    trim=18 18 18 22, clip
  ]{\detokenize{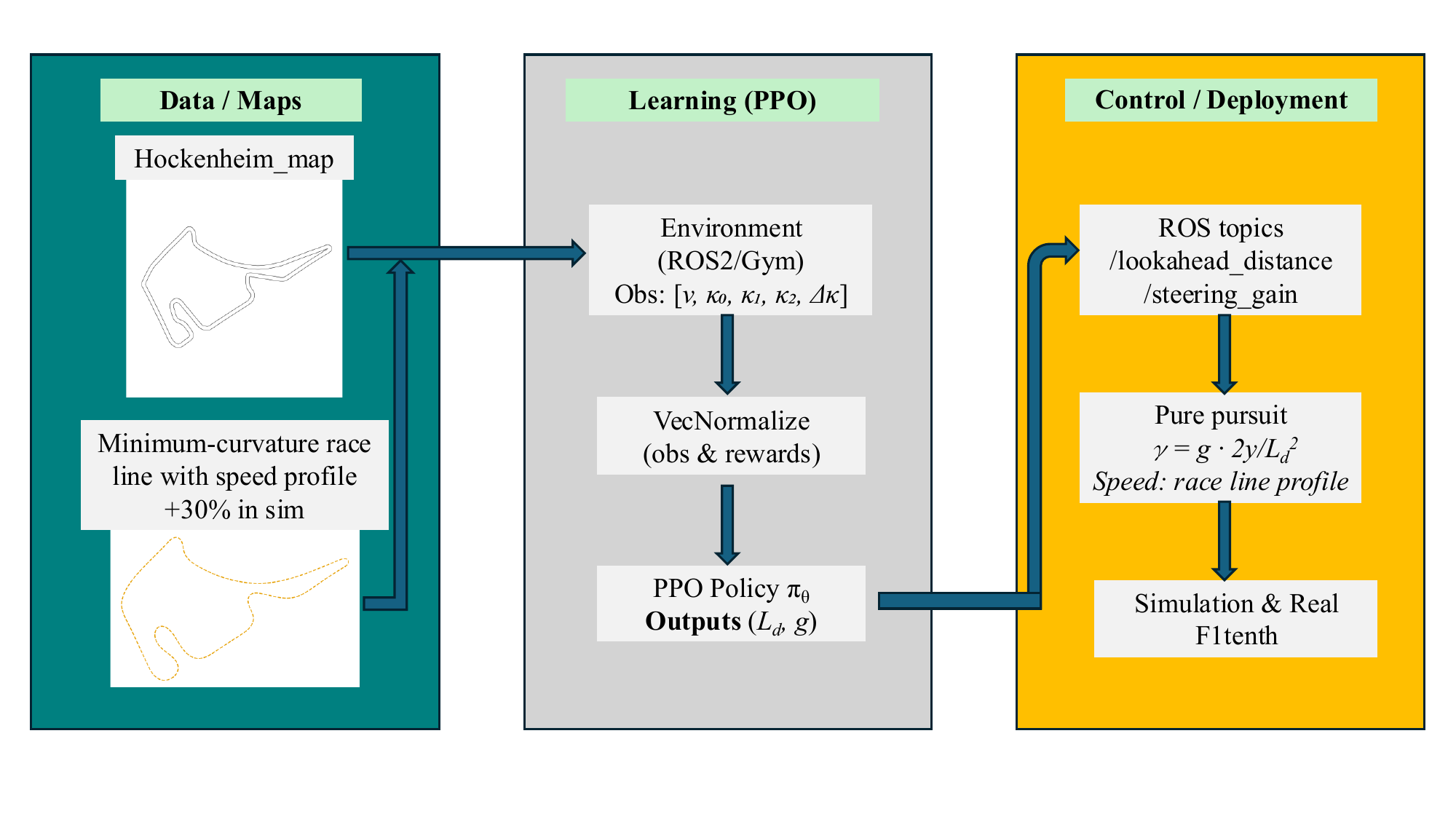}}

  \caption{Pipeline: mapping and minimum-curvature raceline feed the simulator and a baseline PP controller. A PPO agent learns $(L_d,g)$ and integrates with PP via ROS topics. We validate in simulation and deploy on a real F1TENTH car.}
  \label{fig:overview}
  %\vspace{-6pt} % optional; keep or adjust if you need space
\end{figure}

\subsection{System Overview and Experimental Setup}
\Cref{fig:overview} summarizes the components.
Using the F1tenth \textit{Hockenheim} map, we load a raceline with curvature and a reference speed profile.
The raceline drives both the \textbf{Environment} and a baseline \textbf{PP} controller.
Rollouts train a \textbf{PPO} agent that outputs $(L_d, g)$ each step, which are lightly smoothed and published on ROS topics; PP then computes steering and speed commands.

\subsection{Observation and Action Spaces}
At time $t$ the agent observes
\begin{equation}
s_t=\begin{bmatrix} v_t \\ \kappa_{0,t} \\ \kappa_{1,t} \\ \kappa_{2,t} \\ \Delta\kappa_t \end{bmatrix},
\end{equation}
where $v_t$ is speed, $\kappa_{0,1,2}$ are absolute curvatures at near/mid/far horizon taps along the raceline, and $\Delta\kappa_t=\kappa_{1,t}-\kappa_{0,t}$.
The action is a 2-D continuous vector,
\begin{equation}
a_t =
\begin{bmatrix}
L_{t+1}\\[2pt] g_{t+1}
\end{bmatrix},\qquad
\begin{aligned}
L_{t+1} &\in [0.35, 4.0]\ \text{m},\\
g_{t+1} &\in [0.45, 1.15].
\end{aligned}
\end{equation}
followed by a first-order exponential smoother (per-component) before publishing:
\begin{equation}
\begin{aligned}
\tilde{L}_{t+1} &= \beta_L L_{t+1} + (1-\beta_L)\tilde{L}_t,\\
\tilde{g}_{t+1} &= \beta_g g_{t+1} + (1-\beta_g)\tilde{g}_t.
\end{aligned}
\end{equation}
with $\beta_L=\beta_g=0.2$.
Conceptually, the policy implements a mapping $s_t \mapsto (L_{t+1}, g_{t+1})$ that uses
current speed and previewed curvature to adjust the PP lookahead distance and steering gain
before they are smoothed and applied.

\noindent\textbf{Curvature preview taps.}
Let the raceline be discretized into $N$ waypoints $\{p_i=(x_i,y_i)\}_{i=0}^{N-1}$ with
associated curvature samples $\{\kappa_i\}_{i=0}^{N-1}$.
At time $t$, we find the nearest waypoint index
$i_t^\star=\arg\min_i \|p_t-p_i\|_2$, where $p_t$ is the current vehicle position.
We then form three fixed preview taps using waypoint-index offsets
$k\in\{0,5,12\}$ (with wrap-around modulo $N$):
$\mathcal{I}_t(k)=(i_t^\star+k)\bmod N$.
The curvature features are the absolute curvatures at these taps,
$\kappa_{j,t}=|\kappa_{\mathcal{I}_t(k_j)}|$ for $k_0{=}0$, $k_1{=}5$, $k_2{=}12$,
and $\Delta\kappa_t=\kappa_{1,t}-\kappa_{0,t}$.
With our waypoint spacing, $k=5$--$12$ corresponds to approximately $1$--$4$\,m of lookahead.

\subsection{Reward Design}
The reward balances speed, agreement with teacher targets, smoothness, curvature exposure, straight behavior, progress, and safety:

\begin{equation}
\small
\begin{aligned}
R_t &= w_v v_t
     - w_L\,\big|\tilde{L}_t - L_t^*\big|
     - w_G\,\big|\tilde{g}_t - g_t^*\big|
     - w_{jL}\,\big|\tilde{L}_t - \tilde{L}_{t-1}\big|\\
     &\quad- w_{jG}\,\big|\tilde{g}_t - \tilde{g}_{t-1}\big|
     - w_k\,|\kappa_t|
     - w_{\times}\,\big(\tilde{L}_t\,\kappa^{\max}_t\big)\\
&\quad + w_{\text{pre}}\,\mathbb{I}_{\text{bend}}(\kappa^{\max}_t)\,
          \mathbb{I}\!\left[\tilde{L}_t\le \ell(v_t)\right]\\
&\quad - w_c\,\mathbb{I}_{\mathrm{collision}}
     - w_s\,\mathbb{I}_{\mathrm{slow}}
     + w_p\,\Delta p_t\,.
\end{aligned}
\end{equation}
where $\kappa^{\max}_t=\max(\kappa_{0,t},\kappa_{1,t},\kappa_{2,t})$, $\kappa_t$ is a smoothed local curvature, and $\Delta p_t$ counts newly passed waypoints.
The teacher targets are
\begin{equation}
\begin{aligned}
L_t^* &= \mathrm{clip}\!\big(0.50 + 0.28\,v_t - 3.5\,\kappa^{\max}_t,\ 0.35, 4.0\big),\\
g_t^* &= \mathrm{clip}\!\big(m v_t + b,\ 0.45, 1.15\big).
\end{aligned}
\end{equation}
with $m=(g_{\min}-g_{\max})/(v_{\max}-v_{\min})$ and $b=g_{\max}-m v_{\min}$ (we use $v_{\min}{=}3$, $v_{\max}{=}18$, $g_{\max}{=}0.9$, $g_{\min}{=}0.65$).
We give a small bonus ($w_{\text{pre}}$) for \emph{pre-shortening} lookahead before pronounced bends, and gently discourage overly large $g$ on straights.
Collisions are detected if the minimum LiDAR range $<0.2$\,m; we also penalize low speed and stalling.
Rewards are clipped $R_t\in[-30,\,100]$ for stability.

\noindent\textbf{Reward weights.}
Table~\ref{tab:reward_weights} lists the weight values used for all PPO training runs.

\begin{table}[t]
\centering
\caption{Reward weights used for PPO training.}
\label{tab:reward_weights}
\setlength{\tabcolsep}{4pt}
\begin{tabular}{l c}
\toprule
Term & Weight \\
\midrule
$w_v$            & 1.8 \\
$w_L$            & 3.0 \\
$w_G$            & 0 \\
$w_{jL}$         & 0.4 \\
$w_{jG}$         & 0 \\
$w_k$            & 1.5 \\
$w_{\times}$     & 2.0 \\
$w_{\text{pre}}$ & 1.5 \\
$w_c$            & 10.0 \\
$w_s$            & 0.5 \\
$w_p$            & 1.0 \\
\bottomrule
\end{tabular}
\end{table}

\subsection{Pure Pursuit Control}
Given the target point at a distance $\ L_d$ along the raceline, transformed to the vehicle frame $(x',y')$, PP computes
\begin{equation}
\gamma = g\cdot \frac{2y'}{L_d^{2}},\qquad
\gamma\in[-0.35,\ 0.35],
\end{equation}
and publishes the steering command (we use $\gamma$ directly as the steering angle command in our setup). The speed command follows the raceline’s reference profile (capped on the real car for safety). If RL commands are stale beyond a short timeout, PP reverts to a linear teacher for $(L_d,g)$ until fresh actions resume.
Here $L_d$ controls the lookahead distance and therefore the trade-off between responsiveness
and smoothness, while $g$ scales the curvature implied by the lookahead geometry and sets how
aggressively the vehicle turns toward the target point.

\paragraph*{Why $L_d$ and $g$ are non-redundant.}
$L_d$ determines the target point (preview geometry), while $g$ scales the resulting curvature command after the target is fixed. Hence $g$ acts as a pure multiplier, whereas $L_d$ also moves the target point:
\(\partial\gamma/\partial g = 2y'/L_d^2\),
whereas
\(\frac{d\gamma}{dL_d}\)
contains both an explicit term
\(-4gy'/L_d^3\)
and an implicit term
\((2g/L_d^2)\,dy'/dL_d\)
due to the \(L_d\)-dependent target point \((x',y')\). Ablations confirm the benefit of jointly adapting $(L_d,g)$.

\subsection{Vehicle Dynamics Model}
We model the platform with the standard kinematic bicycle dynamics:
\begin{align}
\dot{x} &= v\cos\theta, &
\dot{y} &= v\sin\theta, \\
\dot{\theta} &= \frac{v}{L}\tan\delta, &
\dot{v} &= a,
\end{align}
where $(x,y)$ denote the position, $\theta$ the heading, $v$ the longitudinal speed, $a$ the commanded acceleration, $\delta$ the steering input, and $L$ the wheelbase. Under small-slip conditions, this simplified, nonholonomic model provides an accurate description for our high-rate path-tracking tasks in simulation and is sufficiently faithful for transfer to the real F1TENTH platform.

\subsection{Proximal Policy Optimization (PPO)}
We train the stochastic policy $\pi_\theta(a_t \mid s_t)$—which outputs the 2\!-D action $(L_{t+1},\,g_{t+1})$—using Proximal Policy Optimization (PPO) from Stable-Baselines3 \cite{ppo-stable-baselines3}. PPO is a first-order policy-gradient method that constrains each update via a clipped surrogate objective, which helps prevent large, destabilizing policy shifts between iterations.

Our training setup uses on-policy rollouts of length $n_{\text{steps}}=4096$ with a minibatch size of $256$ and $n_{\text{epochs}}=5$ optimization passes per update. We adopt $\gamma=0.99$ for discounting and Generalized Advantage Estimation with $\lambda=0.98$. The probability ratio is clipped with $\epsilon=0.2$, and we monitor a target KL of $0.015$ to guard against overly aggressive updates. We include an entropy bonus (coefficient $0.02$) to encourage exploration and weight the value-function loss with $0.6$. Gradients are clipped at a global norm of $0.7$. The default learning rate follows a linear decay
$\ell(f)=\ell_0 f$ with $\ell_0=2.4\times10^{-4}$.

We also evaluate a cosine schedule (enabled via \texttt{--lr-schedule cosine}), implemented as
\[
\ell_{\cos}(f)=\tfrac{\ell_0}{2}\bigl(1+\cos\!\bigl(\pi(1-f)\bigr)\bigr),
\]
where $f\in[1,0]$ is the remaining training progress provided by SB3.

Observations and returns are normalized online with \textbf{VecNormalize} for stability.

PPO maximizes the clipped surrogate
\begin{IEEEeqnarray}{l}
\mathcal{L}_{\text{clip}}(\theta) = \nonumber\\[2pt]
\quad \mathbb{E}_t\!\Big[\min\!\big(r_t(\theta)\,\hat A_t,\,
\operatorname{clip}(r_t(\theta),1-\epsilon,1+\epsilon)\,\hat A_t\big)\Big]
\IEEEeqnarraynumspace\label{eq:ppo-clip}
\end{IEEEeqnarray}
where $r_t(\theta)=\pi_\theta(a_t\mid s_t)/\pi_{\theta_{\text{old}}}(a_t\mid s_t)$ and $\hat A_t$ denotes the advantage. The full objective combines policy, value, and entropy terms:
\begin{IEEEeqnarray}{l}
\mathcal{L}_{\text{ppo}}(\theta,\phi) = \nonumber\\[2pt]
\quad -\mathcal{L}_{\text{clip}}(\theta)
+ c_v\,\mathbb{E}_t\!\big[(V_\phi(s_t)-\hat R_t)^2\big] \nonumber\\[2pt]
- c_s\,\mathbb{E}_t\!\big[\mathcal{H}(\pi_\theta(\cdot\mid s_t))\big]
\IEEEeqnarraynumspace\label{eq:ppo-final}
\end{IEEEeqnarray}
with $c_v=0.6$ and $c_s=0.02$ in our implementation.
During training we evaluate the policy periodically on a separate evaluation environment that shares (but does not update) the normalization statistics, and we save both best-model checkpoints and intermediate snapshots.

\subsection{Implementation and Training}
Our pipeline is implemented in ROS2 using \texttt{rclpy} and trained with Stable-Baselines3 PPO. The policy outputs a two-dimensional action $(L_{t+1},\,g_{t+1})$ at each control step; both components are lightly filtered and published on ROS topics consumed by the Pure Pursuit (PP) controller. PP then computes steering directly from the geometric relation $\gamma=g\cdot 2y/L^2$ and applies a raceline speed profile (safely capped on hardware). If RL messages become stale, PP falls back to a linear “teacher’’ rule for $(L,g)$ until fresh actions arrive.

Runs comprise up to \textbf{1.2M} environment steps. We log to \textbf{TensorBoard} and evaluate every \textbf{5{,}000} steps on a separate evaluation environment that \emph{shares} (but does not update) the normalization statistics. The best-performing snapshot is saved automatically, and additional checkpoints are written every \textbf{25{,}000} steps to enable recovery and ablation comparisons.

\begin{figure}[t]
  \centering
  \subfloat[\footnotesize\bfseries Approx.\ KL (target 0.015)]{%
    \includegraphics[width=0.48\linewidth]{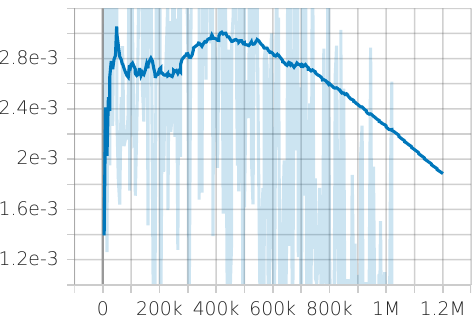}%
  }\hfill
  \subfloat[\footnotesize\bfseries Clip fraction]{%
    \includegraphics[width=0.48\linewidth]{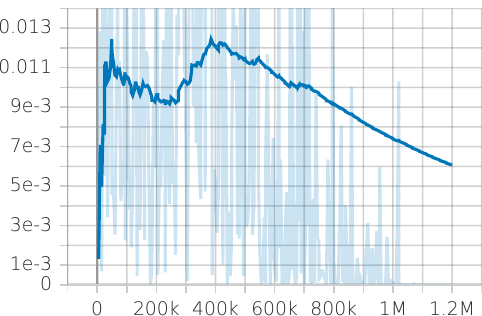}%
  }
  \caption{PPO diagnostics (1/2): update stability metrics. Approx.\ KL and clip fraction remain small after the initial transient, indicating conservative policy updates.}
  \label{fig:ppo_tb_a}
  \vspace{-3mm}
\end{figure}

\begin{figure}[t]
  \centering
  \subfloat[\footnotesize\bfseries Policy action std]{%
    \includegraphics[width=0.48\linewidth]{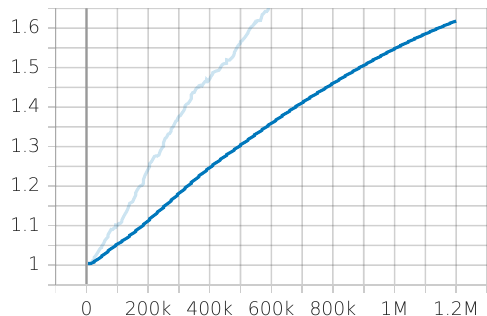}%
  }\hfill
  \subfloat[\footnotesize\bfseries Value loss (critic MSE)]{%
    \includegraphics[width=0.48\linewidth]{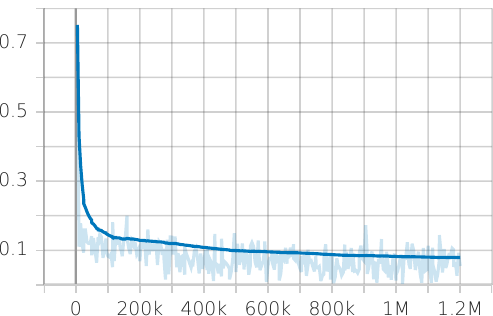}%
  }
  \caption{PPO diagnostics (2/2): exploration and critic learning. The policy action standard deviation decreases as exploration reduces, while the critic loss stabilizes as value estimates improve.}
  \label{fig:ppo_tb_b}
  \vspace{-3mm}
\end{figure}

As training proceeds (Figs.~\ref{fig:ppo_tb_a}, \ref{fig:ppo_tb_b}), the approximate KL and clip fraction remain small after an initial spike and trend downward later in training, the policy’s action standard deviation increases gradually, and the critic’s value loss decreases and stabilizes—together indicating conservative updates with a stabilizing critic.

The full loop executes entirely in simulation (F1TENTH Gym + ROS2), and we repeat runs to check consistency and generalization. The agent emits $(L_d,g)$, which are relayed to PP. PP issues steering/speed commands, and the environment returns the next state and reward. This closed-loop design allows the policy to learn curvature- and speed-aware adjustments for \emph{both} lookahead and steering gain, and it transfers readily to the real vehicle via the same ROS interfaces.

\begin{figure}[!t]
  \centering
  \captionsetup{skip=2pt}

  \subfloat[Fixed \(L\) too small: weaving on a straight.]{%
    \begin{minipage}[b]{0.38\linewidth}
      \centering
      \includegraphics[width=0.6\linewidth, height=1.6in]{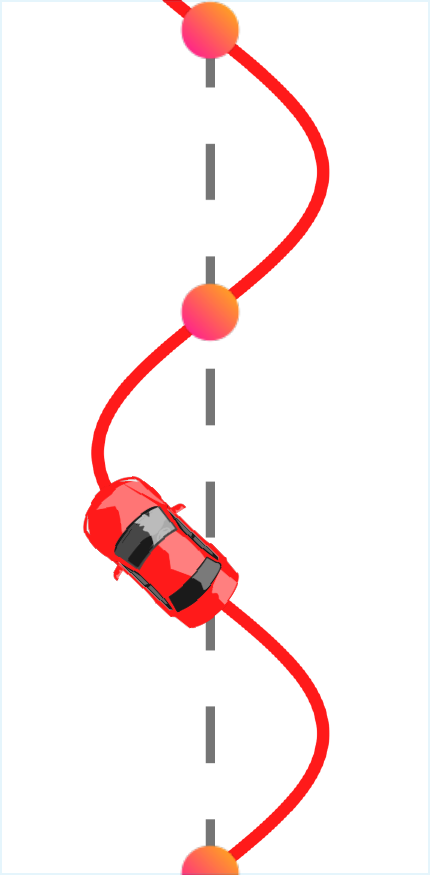}
    \end{minipage}}%
  \hfill
  \subfloat[Fixed \(L\) too large: corner cutting / wall impact.]{%
    \begin{minipage}[b]{0.5\linewidth}
      \centering
      \includegraphics[width=0.92\linewidth]{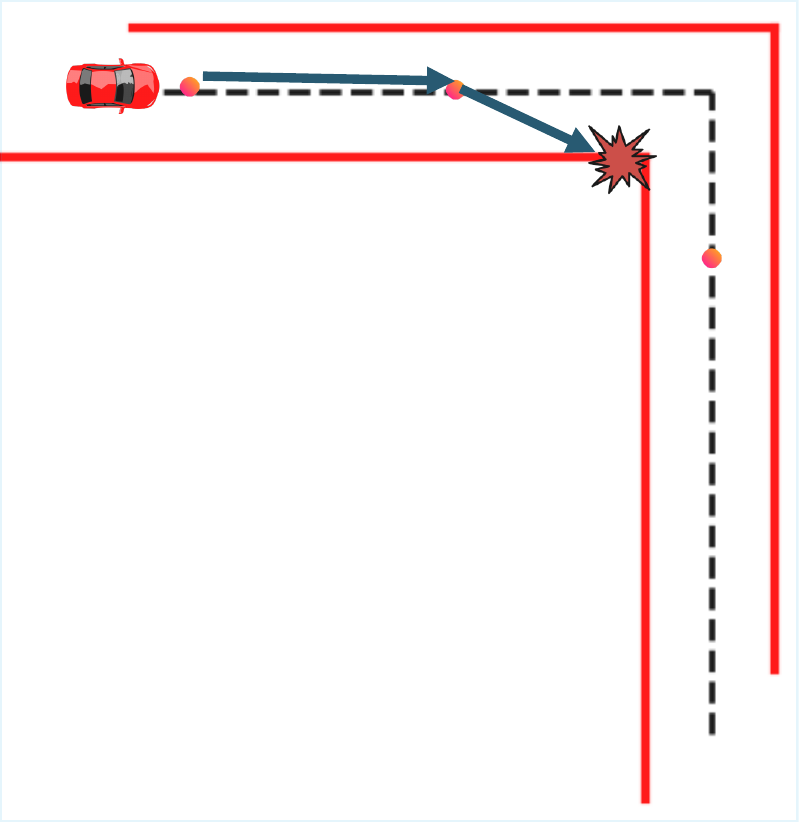}
    \end{minipage}}

  \caption{Classical Pure Pursuit failure modes with fixed lookahead.}
  \label{fig:pp-fail}
  \vspace{-2mm}
\end{figure}

All experiments were executed on a workstation-class desktop (Dell Precision~3660, Intel Core i9\textendash13900, 32\, GiB RAM) running Ubuntu~22.04.5~LTS (64-bit, X11). We observed typical wall-clock training times on the order of a few hours for runs of $\sim$1.2\, M environment steps.

\section{Experimental Results}
\label{sec:experiments}
We evaluate in both simulation and on a physical F1TENTH vehicle. Across all experiments, we compare five controllers:
(i) Fixed PP, (ii) Adaptive PP (linear $v\!\mapsto\!L_d$), (iii) RL--PP ($L_d$ only), (iv) RL--PP (joint $(L_d,g)$), and (v) a kinematic MPC raceline tracker.
All methods track the same minimum-curvature raceline and share the same localization and interfaces for fair comparison.

\par\addvspace{0.7\baselineskip}

\subsection{Baseline: Kinematic MPC}
\label{sec:stronger_baselines}

\paragraph*{Kinematic MPC raceline tracker (baseline).}
As a stronger model-based reference, we implement a kinematic MPC waypoint tracker that follows the same minimum-curvature raceline used by PP. The MPC state is
$\mathbf{x}=[x,\;y,\;v,\;\psi]^\top$ and the control is $\mathbf{u}=[a,\;\delta]^\top$.
We use a horizon of $T_K=8$ with timestep $\Delta t=0.1$~s (0.8~s lookahead). The horizon
reference $\{\mathbf{x}^{\mathrm{ref}}_t\}_{t=0}^{T_K}$ is generated by advancing along the raceline
proportional to the current speed.

At each control step, we solve a QP that penalizes tracking error, control effort, and control-rate changes:
\begin{equation}
\small
\begin{aligned}
\min_{\mathbf{x}_{0:T_K},\,\mathbf{u}_{0:T_K-1}}\;\;
&\textstyle\sum_{t=0}^{T_K-1}\|\mathbf{x}_t-\mathbf{x}^{\mathrm{ref}}_t\|_{Q}^{2}
+ \|\mathbf{x}_{T_K}-\mathbf{x}^{\mathrm{ref}}_{T_K}\|_{Q_f}^{2} \\
&\textstyle+ \sum_{t=0}^{T_K-1}\|\mathbf{u}_t\|_{R}^{2}
+ \sum_{t=0}^{T_K-2}\|\mathbf{u}_{t+1}-\mathbf{u}_{t}\|_{R_\Delta}^{2}.
\end{aligned}
\label{eq:mpc_obj}
\end{equation}

with $Q=Q_f=\mathrm{diag}(13.5,\;13.5,\;5.5,\;13.0)$ for $(x,y,v,\psi)$,
$R=\mathrm{diag}(0.01,\;5.0)$ for $(a,\delta)$, and
$R_\Delta=\mathrm{diag} (0.01,\;5.0)$.

We impose actuator and rate limits consistent with our platform:
\begin{equation}
\small
\begin{aligned}
\delta &\in [-0.4189,\;0.4189]\ \text{rad}, \qquad |a| \le 3.0\ \text{m/s}^2, \\
|\delta_{t+1}-\delta_t| &\le \dot\delta_{\max}\Delta t, \qquad \dot\delta_{\max}=180^\circ/\text{s}.
\end{aligned}
\end{equation}

The QP is solved online using CVXPY with OSQP. \\

\begin{figure*}[t]
  \centering
  \captionsetup{skip=2pt}

  \subfloat[Hockenheim]{%
    \begin{minipage}[b]{0.32\textwidth}
      \centering
      \includegraphics[height=1.6in,keepaspectratio, width= 2in]{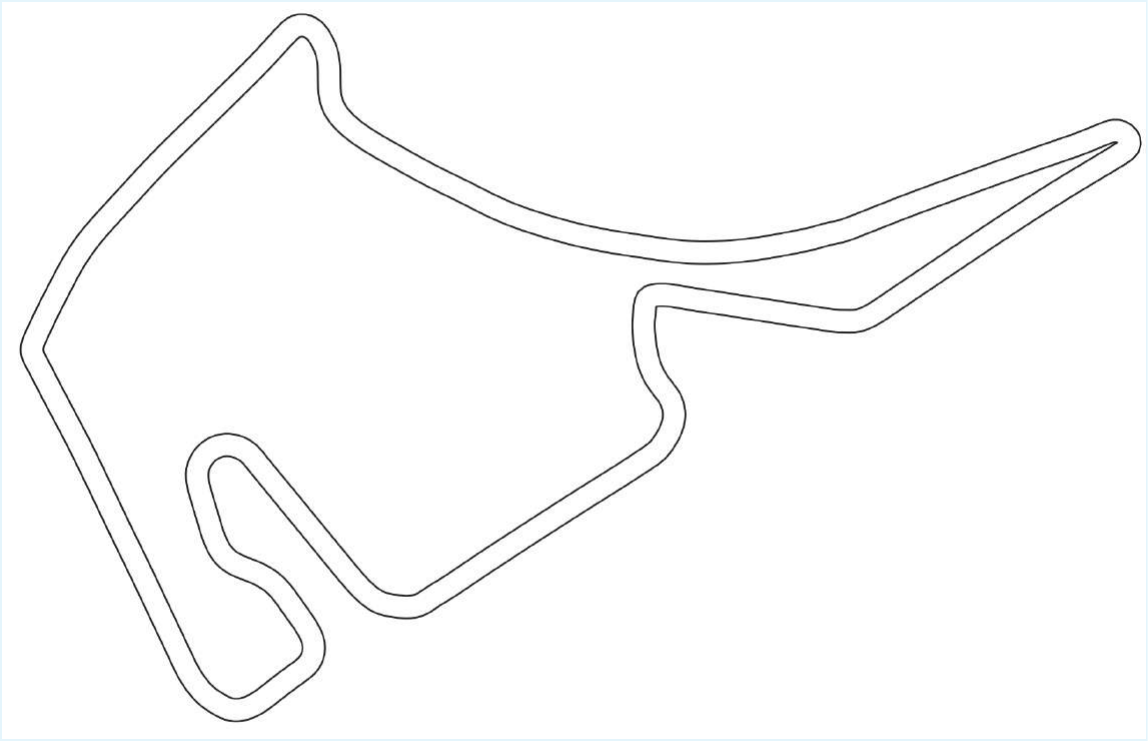}
    \end{minipage}}%
  \hfill
  \subfloat[Montreal]{%
    \begin{minipage}[b]{0.32\textwidth}
      \centering
      \includegraphics[height=1.6in,keepaspectratio,angle=270,origin=c]{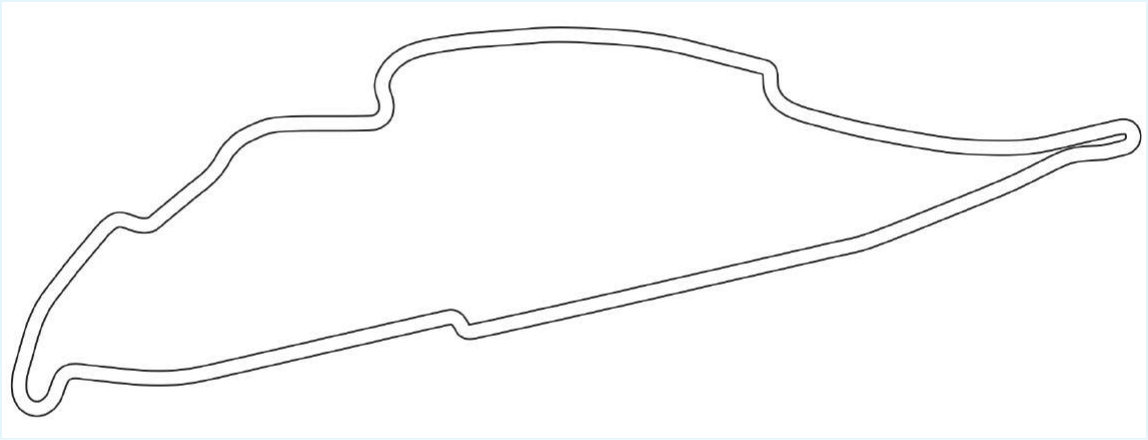}
      % or angle=-90
    \end{minipage}}%
  \hfill
  \subfloat[Yas Marina]{%
    \begin{minipage}[b]{0.32\textwidth}
      \centering
      \includegraphics[height=1.6in,keepaspectratio,angle=270,origin=c]{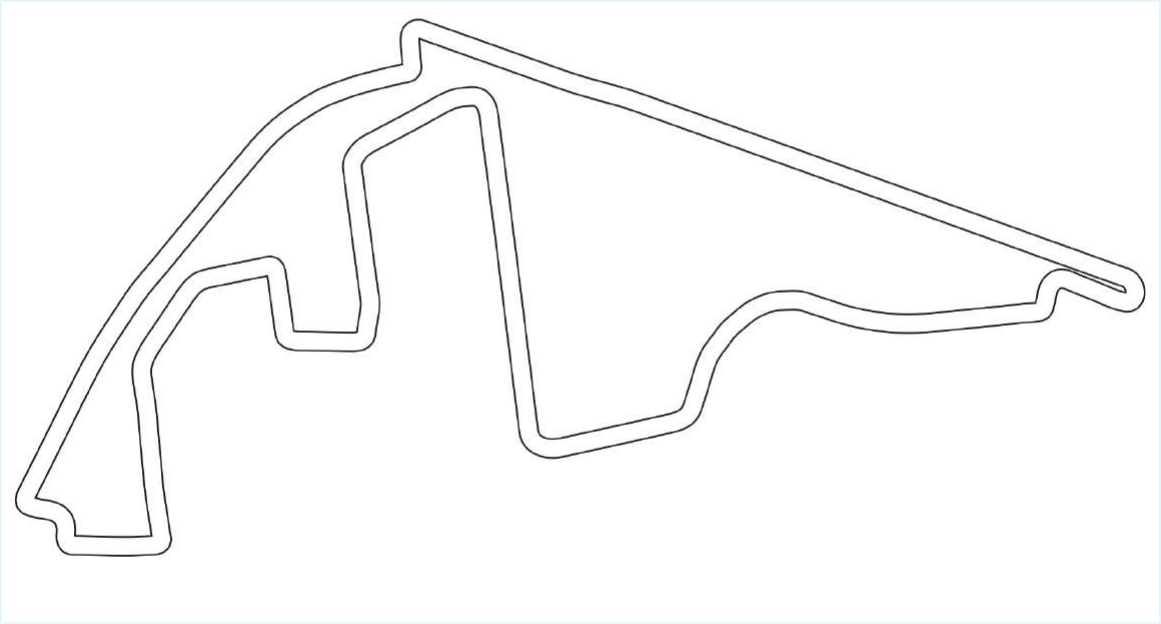}
      % or angle=-90
    \end{minipage}}%

  \caption{Simulator maps used in our study: (a) Hockenheim for training, (b) Montreal, and (c) YasMarina for evaluation.}
  \label{fig:sim-maps}
  \vspace{-4pt}
\end{figure*}

% Qualitative failure zooms (two-column)
\begin{figure*}[t]
  \centering
  \captionsetup{skip=2pt} % keep captions compact to save space
  \subfloat[Montreal]{%
    \includegraphics[width=0.49\textwidth]{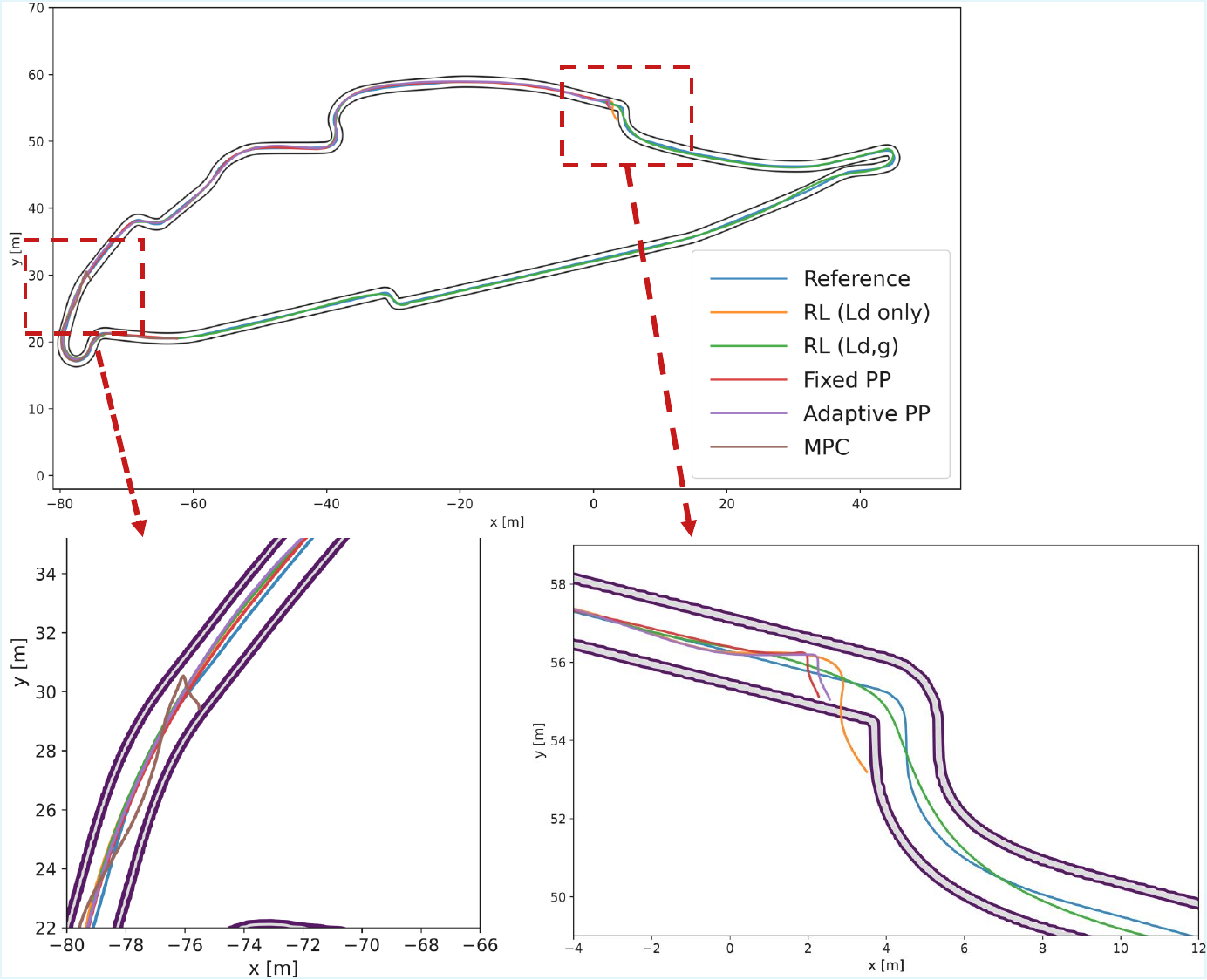}}\hfill
  \subfloat[Yas Marina]{%
    \includegraphics[width=0.49\textwidth, height=2.8 in]{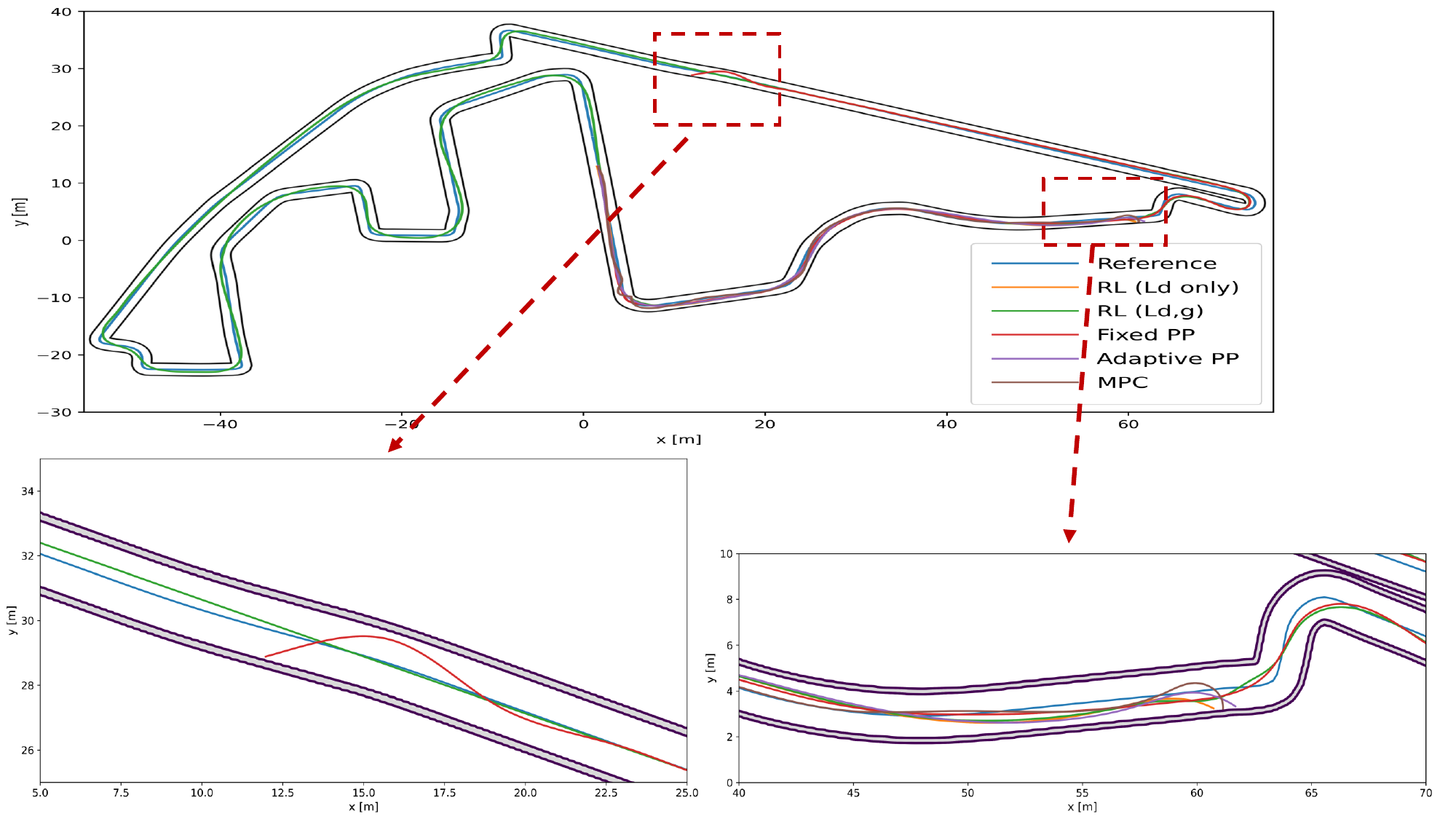}}
  \caption{Qualitative comparison in highlighted regions on two unseen tracks. For each track, the full-layout plot (top) indicates the zoom window shown below. The zoom overlays the reference raceline and all five controllers. The learned RL--PP policies follow the reference closely while remaining within the track boundaries, whereas Fixed PP and Adaptive PP (linear \(v\!\mapsto\!L_d\)) show larger deviations in these regions. The kinematic MPC tracker provides a model-based comparison under the same reference and constraints. Colors follow the legend.}

  \label{fig:qual-zooms}
  \vspace{-4pt}
\end{figure*}

\paragraph*{Fixed-Lookahead Failure Modes.}
Classical Pure Pursuit (PP) with a \emph{fixed} lookahead $L$ is highly sensitive to the track geometry and the chosen speed profile. If $L$ is set too small, the vehicle tends to oscillate on straights and cut inside through bends; if $L$ is set too large, straight-line tracking appears stable, but the car understeers and struggles in tighter corners, as shown in \Cref{fig:pp-fail}. In practice, deploying PP on a new track often requires re-tuning $L$, and performance can remain sensitive despite this effort. In our setting, these limitations motivate learning to adapt not only $L$ but also the \emph{steering gain} $g$ online.

\par\addvspace{0.7\baselineskip}

\begin{figure}[t]
  \centering
  \includegraphics[width=\columnwidth,trim=2cm 1.5cm 2cm 1.5cm,clip]{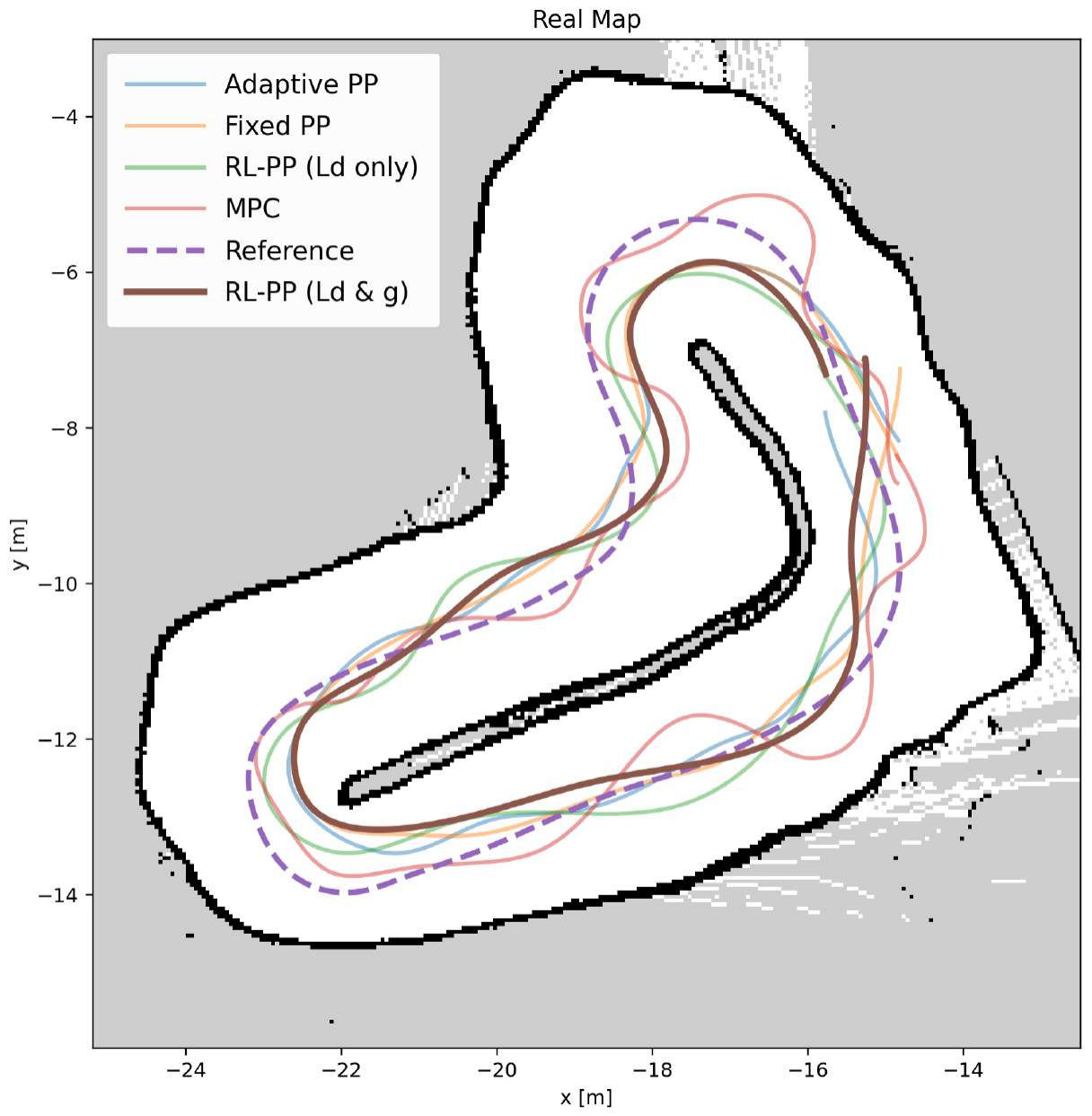}
  \caption{Real-car tracking comparison using a shared raceline speed profile capped at \(v_{\max}=6~\mathrm{m/s}\): overlaid trajectories for all five controllers against the same reference raceline. Using the same capped profile emphasizes differences in lateral tracking and stability while keeping longitudinal commands comparable across controllers.}

  \label{fig:real-overlays}
\end{figure}

\paragraph*{Challenges}
Although our hardware experiments reach \(v_{\max}=6~\mathrm{m/s}\), they still operate below the most aggressive dynamics explored in simulation (via speed-profile scaling). At higher speeds, tire slip and unmodeled
nonlinearities become more prominent; thus we use the high-speed simulation stress tests to complement the
real-car validation while maintaining safety on the physical track.

\paragraph*{Setup and zero-shot evaluation.}
We train a PPO policy that outputs \textbf{both} $(L,g)$ entirely in simulation on the \textit{Hockenheim} F1TENTH racetrack, then evaluate \emph{without any finetuning} on two unseen layouts: \textit{Montreal} and \textit{Yas Marina} from the \texttt{f1tenth\_racetracks} collection \cite{f1tenth_racetracks}. To increase task difficulty and encourage generalization, we \emph{simulate} training with the raceline speed profile uniformly scaled by \textbf{+30\%}, which yields a maximum reference speed of \(\mathbf{15.6}\,\mathrm{m/s}\) and a minimum of \(\mathbf{4.16}\,\mathrm{m/s}\).

\noindent\textit{Track selection.} Training uses \emph{Hockenheim}; \emph{Montreal} and \emph{Yas Marina} are held out for zero-shot tests (Fig.~\ref{fig:sim-maps}). 
\par\addvspace{0.7\baselineskip}

\paragraph*{Safety-teacher activation rate.}
To confirm that results are driven by the learned policy rather than the linear fallback,
we log whether the controller uses PPO outputs (\emph{RL mode}) or the linear teacher (\emph{teacher mode})
at each control step (teacher mode triggers only if RL outputs are unavailable/stale beyond the timeout).
Across our evaluation runs, the teacher was never triggered on either track:
Montreal \(0/8261\) steps (0.000\%), \(0\) events; Yas Marina \(0/11228\) steps (0.000\%), \(0\) events
(max continuous teacher duration \(0.000\,\mathrm{s}\)).

\paragraph*{Qualitative robustness.}
\Cref{fig:qual-zooms} compares representative challenging regions on \emph{Montreal} and
\emph{Yas Marina}. For each track, the full-track plot marks the area enlarged below, where we
overlay the reference raceline and all five controllers. Across both tracks, the learned RL--PP policies
follow the reference closely through corner entry, mid-corner, and exit while staying within the track
boundaries; the joint $(L_d,g)$ variant is typically the closest match, while the $L_d$-only policy shows slightly larger offsets in the highest-curvature parts of the turn. In contrast, fixed-lookahead PP and the velocity-linear adaptive PP exhibit larger geometry-dependent deviations in the highlighted regions (e.g., earlier inside cutting or larger exit drift), consistent with their reliance on a single hand-designed scheduling rule. The kinematic MPC tracker follows the same reference and provides a
model-based comparison; under our horizon and constraints it shows modest rounding/offset in the
highlighted regions. Overall, these qualitative observations align with the lap-time results: adapting
\emph{both} lookahead and steering gain improves robustness relative to fixed-parameter and single-rule
baselines.

\par\addvspace{0.7\baselineskip}
% Two-column maps figure (IEEEtran)

\subsection{Ablation Study: Benefit of Joint $(L_d,g)$ Tuning}
\label{sec:ablation}
A key design choice in our framework is to adapt \emph{both} the Pure Pursuit lookahead
distance $L_d$ and the steering gain $g$. To isolate the value of learning $g$ in addition
to $L_d$, we compare two learned variants trained under the same PPO setup and observation
space:

\begin{itemize}
\item \textbf{RL--PP (joint)}: the policy outputs $(L_d, g)$.
\item \textbf{RL--PP ($L_d$ only)}: the policy outputs $L_d$ while $g$ is held fixed at a constant
$g=g_0$.
\end{itemize}

\noindent\textit{Choice of $g_0$.}
We set $g_0$ to the best-performing fixed gain selected by validation in simulation
(i.e., a short sweep over a small grid of constant $g$ values, evaluated under the same
completion criterion used in our main results). This ensures the $L_d$-only ablation is not
artificially weakened by a poor fixed gain.

Both variants are trained on \emph{Hockenheim} and evaluated zero-shot on \emph{Montreal} and
\emph{Yas Marina} under the same protocol as the main results. The results are reported in
Tables~\ref{tab:montreal-times} and~\ref{tab:yas-times} (rows \textbf{RL--PP (joint)} and
\textbf{RL--PP ($L_d$ only)}), and in the real-car evaluation in Table~\ref{tab:realcar-times}.

Overall, the joint policy achieves the best combination of lap time and stability, indicating
that learning $g$ in addition to $L_d$ provides measurable benefit beyond adaptive lookahead alone.

\begin{figure}[t]
  \centering
  \subfloat[Schedule vs.\ curvature/speed.]{%
    \includegraphics[width=\linewidth]{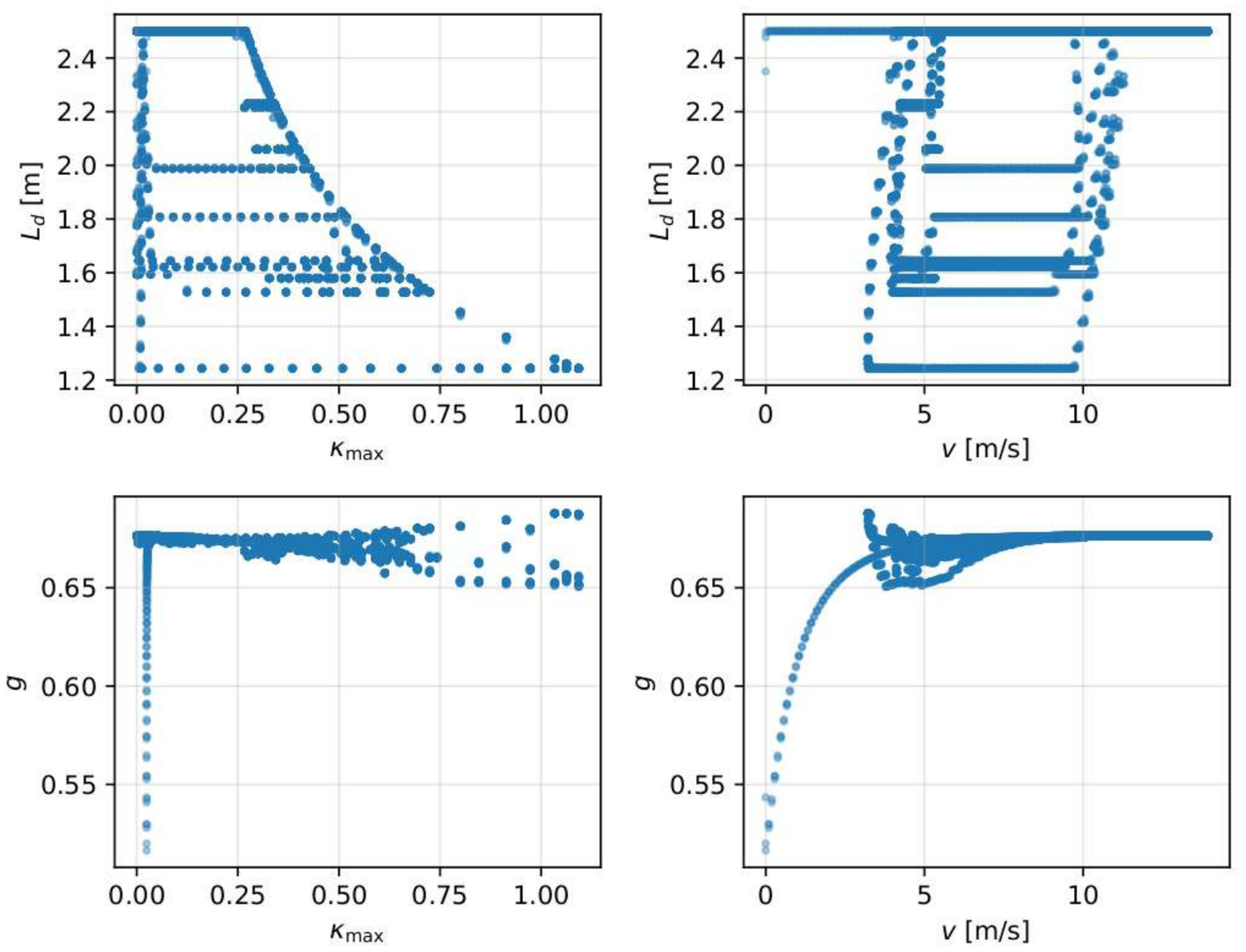}%
    \label{fig:policy_maps}
  }\\[1mm]
  \subfloat[Along-lap evolution.]{%
    \includegraphics[width=\linewidth]{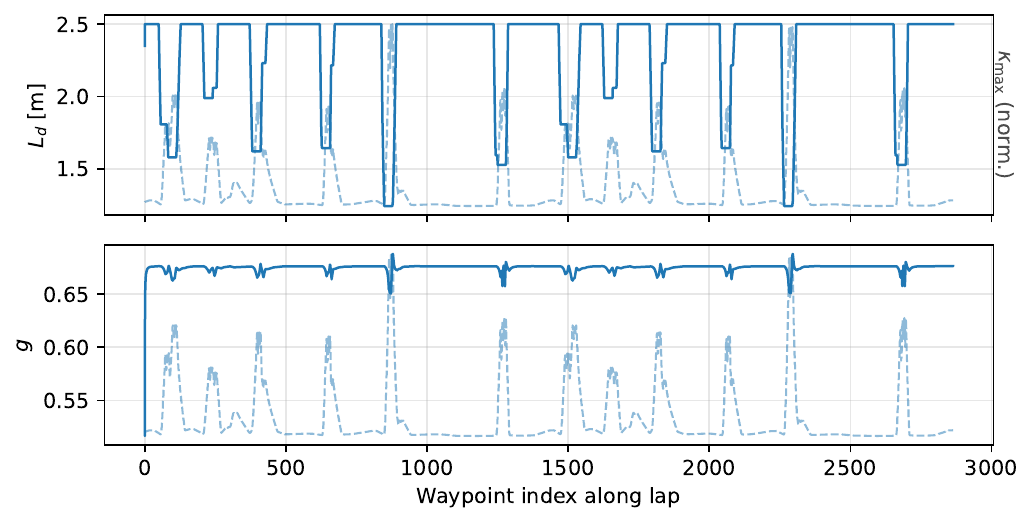}%
    \label{fig:policy_evolution}
  }
  \caption{Policy interpretability on Montreal. (a) Parameter schedule versus curvature and speed along the raceline.
(b) Along-lap evolution of $L_d$ and $g$ plotted against waypoint index, with normalized $\kappa_{\max}$ (dashed)
highlighting high-curvature segments.}

  \label{fig:policy_interpret}
  \vspace{-2mm}
\end{figure}

\paragraph*{Policy interpretability.}
To understand what the learned policy encodes, we log $(L_d,g)$ along the raceline and correlate them
with curvature and speed. Empirically, the policy reduces $L_d$ in high-curvature segments (tighter preview)
and increases $L_d$ on straights. The gain $g$ varies more mildly, remaining near a narrow range while adjusting
slightly with context. We provide plots of $(L_d,g)$ versus $\kappa_t^{\max}$ and $v_t$ in Fig.~\ref{fig:policy_interpret}\subref{fig:policy_maps}, and show the along-lap evolution of $(L_d,g)$ in
Fig.~\ref{fig:policy_interpret}\subref{fig:policy_evolution}.

\paragraph*{Montreal F1TENTH racetrack: quantitative results (10 laps).}
We evaluate five controllers on \emph{Montreal} over 10 consecutive laps. To compare methods under a
common completion criterion, each controller uses the best speed-profile multiplier that achieves
10/10 completed laps in our setup: RL--PP (joint $(L_d,g)$) at \(+16\%\), RL--PP ($L_d$ only) at \(+13\%\),
Adaptive PP (linear \(v\!\mapsto\!L_d\)) at \(+10\%\), Fixed PP at \(-10\%\), and the MPC raceline tracker at
\(-15\%\). The resulting lap-time statistics are summarized in Table~\ref{tab:montreal-times}.

\begin{table}[t]
  \centering
  \caption{Montreal F1TENTH racetrack: lap times over 10 consecutive laps.
  For the 10/10 completion objective, each controller uses the best speed multiplier that yields 10/10 laps in our setup.
  \emph{Adaptive PP (linear)} applies a linear \(v\!\mapsto\!L_d\) rule with bounds \(L_d\in[1.0,2.5]~\mathrm{m}\).}
  \label{tab:montreal-times}
  \begin{tabular*}{\linewidth}{@{\extracolsep{\fill}}l
      S[table-format=2.2]
      S[table-format=1.2]
      S[table-format=2.2]
      S[table-format=2.2]@{}}
    \toprule
    Controller & {Mean} & {Std} & {Min} & {Max} \\
    \midrule
    RL--PP (joint $L_d,g$, \(+16\%\))            & \textbf{32.85} & 0.24 & 32.18 & 32.88 \\
    RL--PP ($L_d$ only, \(+13\%\))               & 33.23          & 0.13 & 33.05 & 33.49 \\
    Adaptive PP (linear $v\!\mapsto\!L_d$, \(+10\%\)) & 34.25     & 0.04 & 34.18 & 34.34 \\
    Fixed PP ($L_d$ fixed, \(-10\%\))            & 41.35    & 0.09 & 41.24 & 41.57 \\
    MPC raceline tracker (\(-15\%\))             & 48.06    & 0.08 & 47.96 & 48.25 \\
    \bottomrule
  \end{tabular*}
\end{table}

%Table title should be on top of the table

From Table~\ref{tab:montreal-times}, RL--PP (joint $(L_d,g)$) achieves a mean lap time of
\(32.85\pm0.24\)~s over ten laps at the \(+16\%\) profile (range \(32.18\)–\(32.88\)~s). The RL--PP
($L_d$ only) variant at \(+13\%\) records \(33.23\pm0.13\)~s (range \(33.05\)–\(33.49\)~s), and the
Adaptive PP (linear \(v\!\mapsto\!L_d\)) baseline at \(+10\%\) achieves \(34.25\pm0.04\)~s (range
\(34.18\)–\(34.34\)~s). Fixed PP requires a reduced profile (\(-10\%\)) to complete all laps and yields
\(41.35\pm0.09\)~s. The MPC raceline tracker attains 10/10 completion at \(-15\%\) with a mean time of
\(48.06\pm0.08\)~s. Overall, the joint RL--PP variant provides the fastest times under our completion
criterion, improving over the linear adaptive baseline while maintaining low lap-to-lap variability.

\par\addvspace{0.7\baselineskip}

\paragraph*{Yas Marina F1TENTH racetrack: quantitative results (10 laps).}
We apply the same protocol as in Montreal: each controller uses the best speed-profile multiplier
that achieves 10/10 completed laps in our setup. On \emph{Yas Marina}, RL--PP (joint $(L_d,g)$) runs at
\( +17.2\% \), RL--PP ($L_d$ only) at \( +15\% \), Adaptive PP (linear \(v\!\mapsto\!L_d\)) at \( +15\% \),
Fixed PP uses the original (non-accelerated) profile, and the MPC raceline tracker completes 10/10 laps
at \(-10\%\). The lap-time statistics (seconds) are reported in Table~\ref{tab:yas-times}.

\begin{table}[t]
  \centering
  \caption{Yas Marina F1TENTH racetrack: lap times over 10 consecutive laps.
  For the 10/10 completion objective, each controller uses the best speed multiplier that yields 10/10 laps in our setup.
  \emph{Adaptive PP (linear)} applies a linear \(v\!\mapsto\!L_d\) rule with bounds \(L_d\in[1.0,2.5]~\mathrm{m}\).}
  \label{tab:yas-times}
  \begin{tabularx}{\linewidth}{@{\extracolsep{\fill}}%
    >{\raggedright\arraybackslash}X
    S[table-format=2.2]
    S[table-format=1.2]
    S[table-format=2.2]
    S[table-format=2.2]@{}}
    \toprule
    Controller & {Mean} & {Std} & {Min} & {Max} \\
    \midrule
    RL--PP (joint $L_d,g$, \(+17.2\%\))                 & {\bfseries 45.10} & 0.07 & 44.96 & 45.20 \\
    RL--PP ($L_d$ only, \(+15\%\))                    & 46.04             & 0.10 & 45.93 & 46.21 \\
    Adaptive PP (linear $v\!\mapsto\!L_d$, [+15]\%) & 46.41    & 0.12 & 46.24 & 46.57 \\
    Fixed PP ($L_d$ fixed, orig.)                     & 52.97             & 0.21 & 52.79 & 53.44 \\
    MPC raceline tracker ([-10]\%)              & 64.01        & 0.25 & 63.60 & 64.51 \\
    \bottomrule
  \end{tabularx}
\end{table}

Across ten laps, the RL policy again leads, with a lower mean time and comparably small dispersion relative to the linear dynamic baseline, and a substantial margin over fixed–\(L\). 

\par\addvspace{0.7\baselineskip}

\paragraph*{Real-car validation (raceline speed profiles capped for safety).}
We evaluated the learned controller on a physical 1:10 racecar (F1TENTH class) \cite{f1tenth}
equipped with a VESC-based drivetrain, a 2D LiDAR (UST-10LX class), and onboard IMU/odometry
running Ubuntu~20.04 with ROS~2 Foxy. A SLAM-based map \cite{slam} was first generated for the test track; during experiments the vehicle localized against this map using scan matching and an MCL
particle filter \cite{pf}.
A raceline for the test track was computed offline using a minimum-curvature optimizer \cite{tum}
and converted into a reference speed profile.

%\noindent\textit{Trajectory overlays (Fig.~\ref{fig:real-overlays}).}
%\Rall{To visualize qualitative tracking behavior under a more conservative setting, we additionally run all
%controllers with the same raceline speed profile capped at} \(v_{\max}=6~\mathrm{m/s}\) \Rone{and overlay their realized trajectories against the reference raceline in Fig.~\ref{fig:real-overlays}. Using a shared profile isolates differences in lateral tracking and stability while avoiding confounding effects from controller-specific speed choices.}

A lightweight ROS~2 wrapper supplies the policy with the same observation channels as in simulation
(speed, curvature along the raceline, and heading error to the lookahead point) and returns the actions
\((L_d, g)\), preserving observation/action parity with the simulator and reducing Sim-to-Real drift due
to interface mismatch.

\noindent\textit{Lap-time evaluation (Table~\ref{tab:realcar-times}).}
All controllers track the same raceline speed profile capped at \(v_{\max}=6~\mathrm{m/s}\) for safety and
a fair comparison. Table~\ref{tab:realcar-times} summarizes the resulting lap times; RL--PP (joint
\(L_d,g\)) achieves the best mean performance (\(9.46\pm0.23\)~s; min \(9.09\), max \(9.82\)), followed by
RL--PP (\(L_d\) only) and Adaptive PP (linear \(v\!\mapsto\!L_d\)). Fixed PP is slightly slower, while the
MPC raceline tracker is substantially slower in our hardware setting.

\begin{table}[h]
  \centering
  \caption{Real-car lap times using a raceline speed profile uniformly capped at \(v_{\max}=6~\mathrm{m/s}\).}

  \label{tab:realcar-times}
  \setlength{\tabcolsep}{6pt}
  \begin{tabular}{lcccc}
    \toprule
    Controller & Mean & Std & Min & Max \\
    \midrule
    RL--PP (joint $L_d,g$) & \textbf{9.46} & 0.23 & 9.09 & 9.82 \\
    RL--PP ($L_d$ only)    & 9.61 & 0.58 & 8.94 & 10.51 \\
    Adaptive PP (linear $v\!\mapsto\!L_d$) & 9.72 & 0.27 & 9.34 & 10.40 \\
    Fixed PP ($L_d$ fixed) & 9.85 & 0.43 & 9.32 & 10.55 \\
    MPC raceline tracker   & 15.42 & 0.47 & 14.48 & 16.17 \\
    \bottomrule
  \end{tabular}
\end{table}

\noindent\textit{Trajectory overlays (Fig.~\ref{fig:real-overlays}).}
To visualize tracking behavior under a more conservative cap, we generate the overlays in
Fig.~\ref{fig:real-overlays} using the same raceline speed profile capped at \(v_{\max}=6~\mathrm{m/s}\).
All five controllers are overlaid against the reference raceline, highlighting that the learned
policies follow the reference smoothly with small lateral deviation.

\section{Conclusion}
We presented a hybrid Pure Pursuit (PP) controller in which a PPO policy \emph{jointly} selects the
lookahead \(L_d\) and steering gain \(g\) from curvature--speed features. The design preserves PP's
geometric steering law and adds only lightweight interfaces (two ROS topics) with a linear fallback,
yielding a deployable and interpretable module.

Across simulation tracks, the RL-tuned PP consistently improves lap time, path-tracking accuracy, and
steering smoothness over both fixed-\(L\), velocity-linear baselines, and MPC raceline tracker (see
Tables~\ref{tab:montreal-times} and \ref{tab:yas-times}). On hardware with a capped raceline speed
profile, RL--PP (joint \((L_d,g)\)) achieves the best mean lap time among the evaluated controllers
(Table~\ref{tab:realcar-times}) and exhibits smooth tracking behavior in the qualitative zooms and
trajectory overlays (Figs.~\ref{fig:real-overlays}).

Overall, adaptive selection of \((L_d,g)\) reduces per-track retuning and transfers from simulation to the real car, though performance depends on a high-quality global reference raceline/localization and may degrade under extreme dynamics, motivating future work on robustness to imperfect references and highly dynamic scenarios.

\balance
\bibliographystyle{IEEEtran}
%\IEEEtriggeratref{22} % <-- change the number until the refs fit in 8 pages
\bibliography{refs}
\end{document}